\documentclass[conference, letterpaper]{IEEEtran}
\usepackage{listings}

\ifCLASSINFOpdf
\else
\fi
\ifCLASSINFOpdf
   \usepackage[pdftex]{graphicx}
\else
\fi

\usepackage[cmex10]{amsmath}
\usepackage{color}
\usepackage{fancyhdr}
\usepackage[caption=false,font=footnotesize]{subfig}
 \usepackage{hyperref}

\renewcommand{\thispagestyle}[2]{} 

\fancypagestyle{plain}{
        \fancyhead{}
        \fancyhead[C]{first page center header}
        \fancyfoot{}
        \fancyfoot[C]{first page center footer}
}
\begin{document}

%
\title{Real Time Offside Detection using a Single Camera in Soccer
}

\author{\IEEEauthorblockN{Shounak Desai}
\IEEEauthorblockA{Department of Computer Science\\Golisano College of Computing and Information Sciences\\
Rochester Institute of Technology\\
sd9649@rit.edu}}


%


\maketitle




%
\IEEEpeerreviewmaketitle

\section*{Abstract}

Technological advancements in soccer have surged over the past decade, transforming aspects of the sport. Unlike binary rules, many soccer regulations, such as the "Offside Rule," rely on subjective interpretation rather than straightforward True or False criteria. The on-field referee holds ultimate authority in adjudicating these nuanced decisions. A significant breakthrough in soccer officiating is the Video Assistant Referee (VAR) system, leveraging a network of 20-30 cameras within stadiums to minimize human errors. VAR's operational scope typically encompasses 10-30 cameras, ensuring high decision accuracy but at a substantial cost. This report proposes an innovative approach to offside detection using a single camera, such as the broadcasting camera, to mitigate expenses associated with sophisticated technological setups.

\section{Introduction}
 
The objective of this research project is to detail the implementation of offside detection using a single camera, specifically the broadcast camera commonly used in soccer matches. This system leverages various image processing techniques, including segmentation, line detection, and 3D-to-2D space interpretation, to achieve accurate offside detection. \\

The background section provides a comprehensive overview of prior implementations related to offside detection using similar techniques. It also discusses the inspiration behind this project, highlighting the need for innovative approaches in soccer technology.
In the Experiment section, a step-by-step explanation will be provided regarding the implementation process. This includes the methodology adopted for image processing, algorithmic decisions, and technical considerations for utilizing a single camera setup.
The Results section presents the anticipated outputs of this project, showcasing the effectiveness and reliability of the offside detection system. Visual representations and statistical analyses will be included to support the findings.
Furthermore, this report discusses the potential impact of this project on the realm of soccer technology. It highlights how advancements in offside detection can contribute to fairer and more accurate officiating.
Additionally, it outlines areas for future research and development to enhance the capabilities of such systems.
To conclude, this report sets the stage for a detailed exploration of offside detection using a single camera, emphasizing its importance in advancing soccer technology and paving the way for future innovations in sports officiating.

\section{Background}

\subsection{What is Offside in Soccer ?}In soccer, a player is offside if they are closer to the opponent's goal line than the ball and the last defender when the ball is passed to them. Being offside means they can't immediately play the ball, preventing unfair advantages. It's like ensuring players don't camp near the goal for easy goals, promoting fair and active gameplay.

\begin{figure}[htbp]
    \centering
    \includegraphics[width=0.4\textwidth]{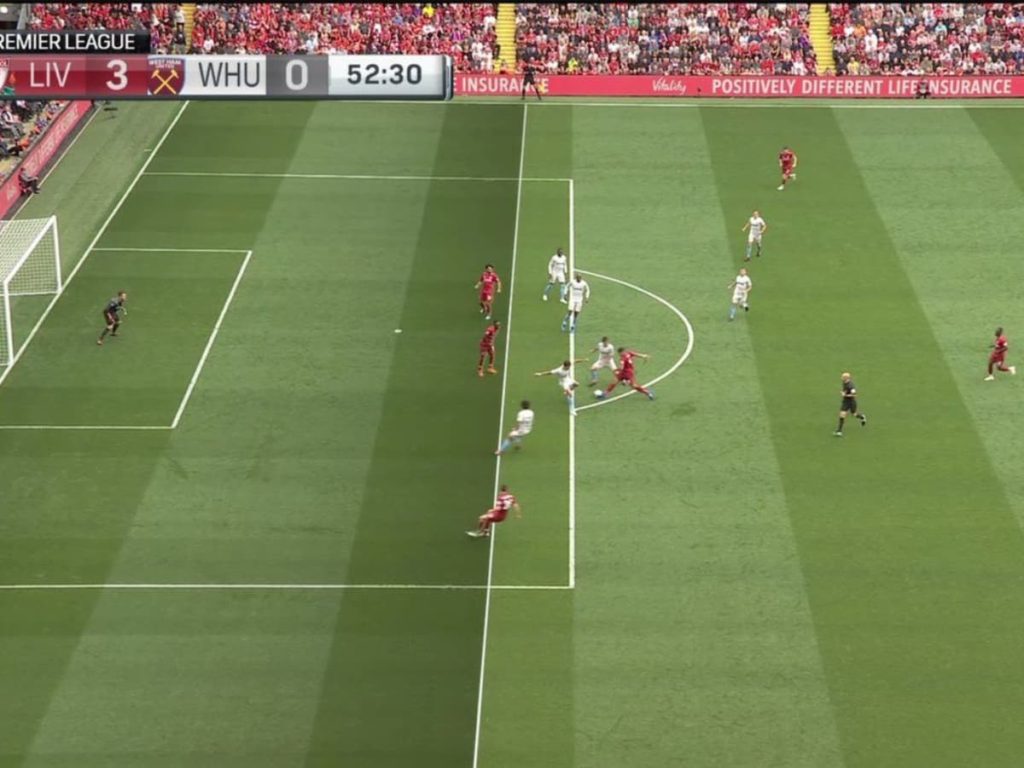}
    \caption{Offside example}
    \label{fig:gan_before_MC}
\end{figure}

In Fig. 1, the players in white jerseys represent the defending team, while those in red jerseys are the attacking team moving from right to left. A virtual line is drawn from the feet of the last defender (the leftmost player in white). This line, called the Offside line, marks the boundary beyond which no player from the attacking team (red jerseys) should be positioned when receiving the ball. If, as depicted in Figure 1, there are three players ahead of this virtual line when the ball is played to them, it constitutes a foul. The game stops, and possession of the ball is awarded to the opposing team.\\

\subsection{Related Work}There has been a lot of similar research already in this field. With the introduction of VAR which uses different cameras at different angles to draw parallel lines is one of the advanced and state-of-the-art tech. I came up with a novel idea to implement the Offside detection. I have majorly referred two papers while working on this project and became my inspiration for the new idea. Vision based offline Vision Based Dynamic Offside Line Marker for Soccer Games; \cite{muthuraman2018vision} explains the step by step implementation of using Image processing techniques to detect lines on the Soccer pitch and then drawing parallel lines in 3D space. Although, this implementation is based on the FIFA Game dataset which is a kind of simulation dataset. Along with this, the authors used the players "Bodies" converting into blobs to draw parallel lines. Since the offside rule is very strict about which body parts are considered as "Offside", this paper does not use those specific body parts. Computer Vision based Offside Detection in soccer; \cite{10.1145/3422844.3423055}, uses the Keypose estimation to detect the only relevant body parts that are used in the Offside rule. This paper implements the offside detection on the body parts of the players on a real dataset for Images. These 2 papers inspired me to come up with a new idea of doing the same Offside detection on Real Video dataset of a broadcasted game and implementing keypose detection to make it accurate.

\section{Experiment}
\subsection{Play Area Segmentation:}  The data used for this project is basically a broadcasted game "Manchester City vs West Ham 2023; Premiere League". The game that we see on TV uses a single static camera which never changes the position and pans left to right or right to left as the game goes on. Fig. 2 shows how one of the frames looks like during the game. We can see audience in the background watching the game, advertisement boards and score board set up by the broadcaster. 

\begin{figure}[htbp]
    \centering
    \includegraphics[width=0.4\textwidth]{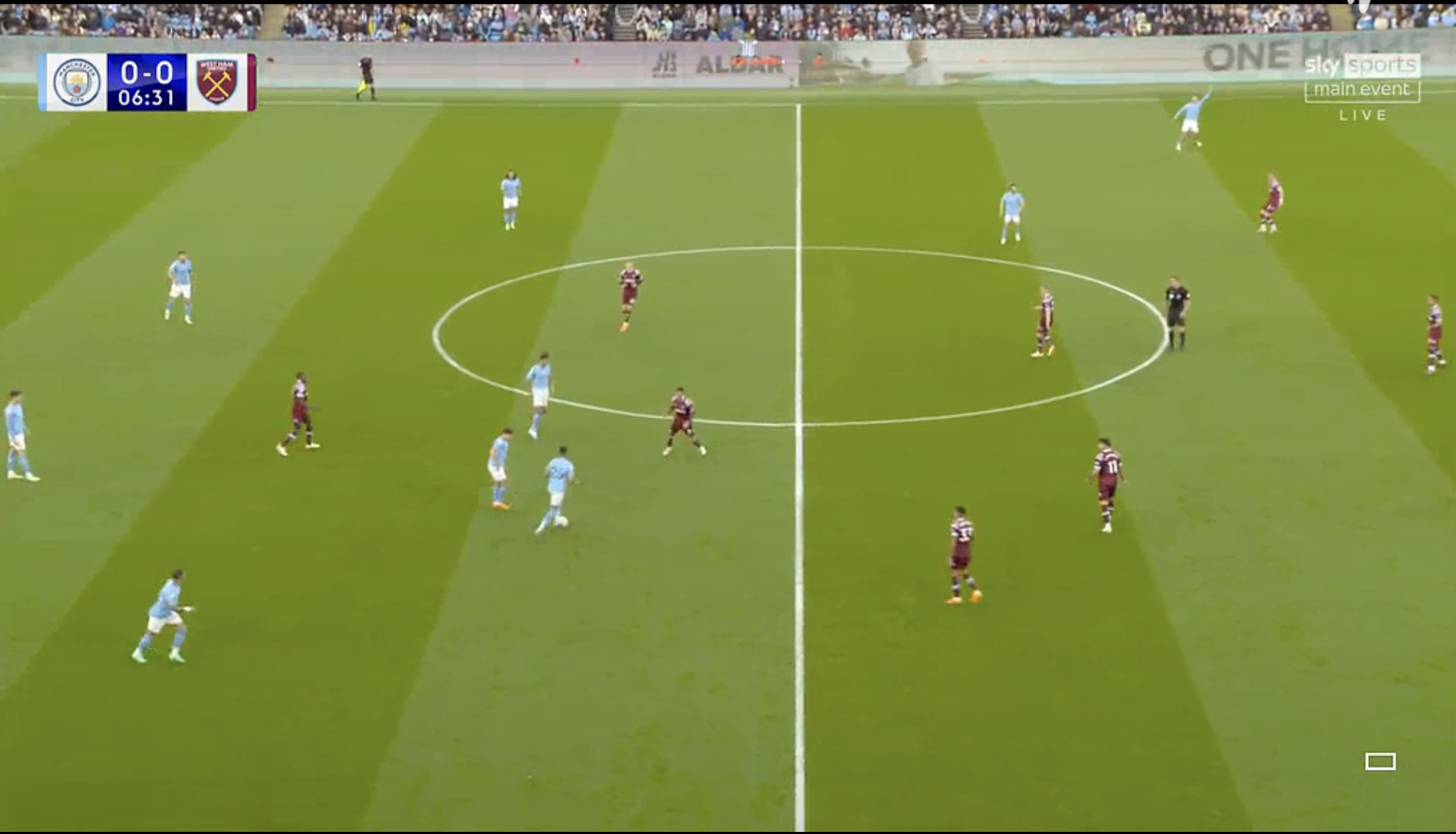}
    \caption{Input Frame/Image}
    \label{fig:gan_before_MC}
\end{figure}

During Image processing we only need the Soccer pitch and the players inside it. Rest of the details become noise in the input and may be responsible for poor image processing. Hence, Segmenting the Playing area becomes a necessary part of the process. Fig. 3. shows the segmented image. I used the OpenCV library to segment the playing area in an image. First, I converted the image from the BGR color space to the HSV color space using the cv.cvtColor() function. Then, I defined a range of green color in HSV format using lower and upper bounds. Next, I created a binary mask by thresholding the HSV image using the cv.inRange() function with the defined green color range.

\begin{figure}[htbp]
    \centering
    \includegraphics[width=0.4\textwidth]{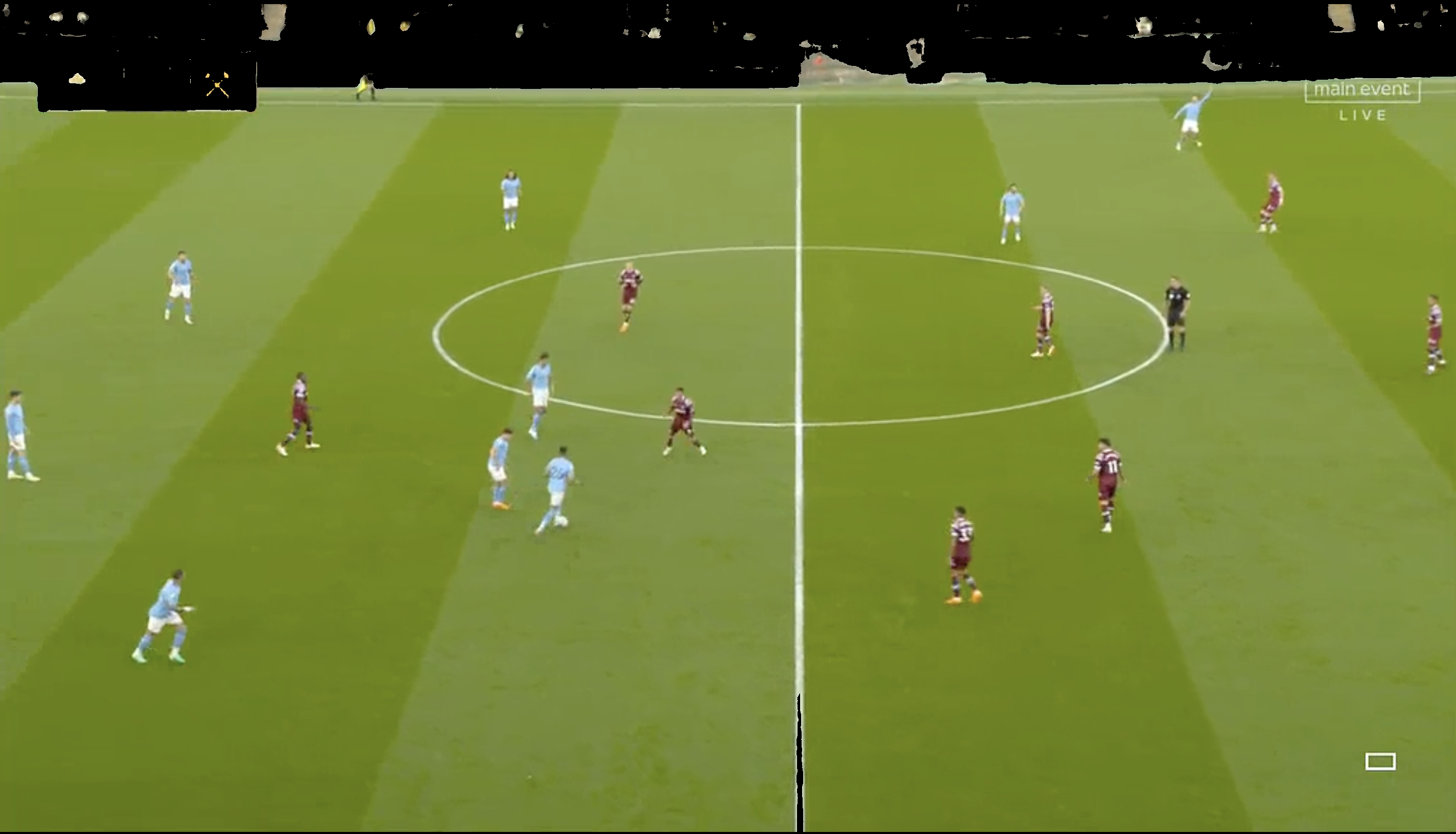}
    \caption{Play Area Segmented Image}
    \label{fig:gan_before_MC}
\end{figure}

After obtaining the binary mask, I found contours in the mask using the cv.findContours() function with the RETR\_EXTERNAL retrieval mode and CHAIN\_APPROX\_SIMPLE contour approximation method. This step helped me identify the boundaries of the green areas in the image.

Using the detected contours, I created a black image with the same dimensions as the original image using np.zeros\_like(). Then, I drew filled contours on this black image using cv.drawContours() to create a mask of the green areas.

Finally, I applied this mask to the original image using bitwise AND operation (cv.bitwise\_and()) to obtain the segmented result, where only the green areas are preserved, and everything else is blacked out as seen in Fig. 3.

Now the image is ready for further processing.

\subsection{Detecting lines on Soccer pitch:}

The light green and dark green patches (check Fig. 2.) was the biggest clue to detect all the parallel lines on the soccer pitch at any point of time in the game. These lines are parallel to the breadth of the field, making all these lines parallel in 3D space. Hence, the idea was to only detect these lines which can be actual references to draw the virtual lines passing through all the players.
To detect these lines, appropriate Image processing is necessary. Hence, i tested different image space CIE-LAB, HSV, BGR etc. and came to the conclusion that the difference in the light green and dark green is maximum when used the Saturation channel from HSV image space. You can see Fig. 4. which shows the difference in light and dark green strips on the field.

\begin{figure}[htbp]
    \centering
    \includegraphics[width=0.4\textwidth]{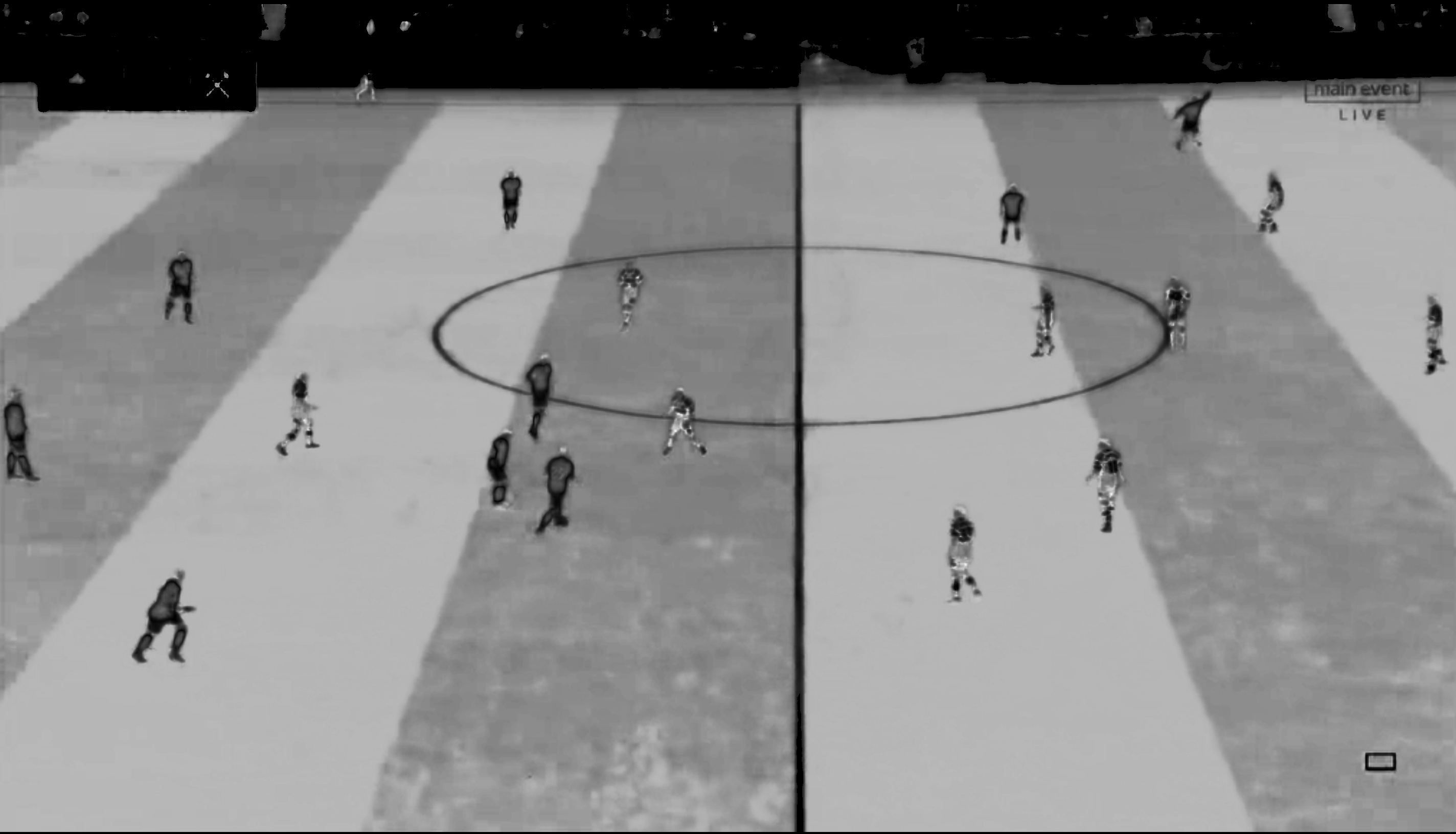}
    \caption{Saturation Channel of HSV space}
    \label{fig:gan_before_MC}
\end{figure}

This makes it easier to find the edges on the field as the gradient is high where the light and dark green strips meet. Hence, using Sobel filter which is the go-to filter for edge detection makes sense. I used the Canny edge detector which is in general a Sobel filter in both x and y directions after using a gaussian filter. After applying canny edge detector we can see strong edges/ lines as expected on the field. See fig. 5. for the output of canny edge detection.

\begin{figure}[htbp]
    \centering
    \includegraphics[width=0.4\textwidth]{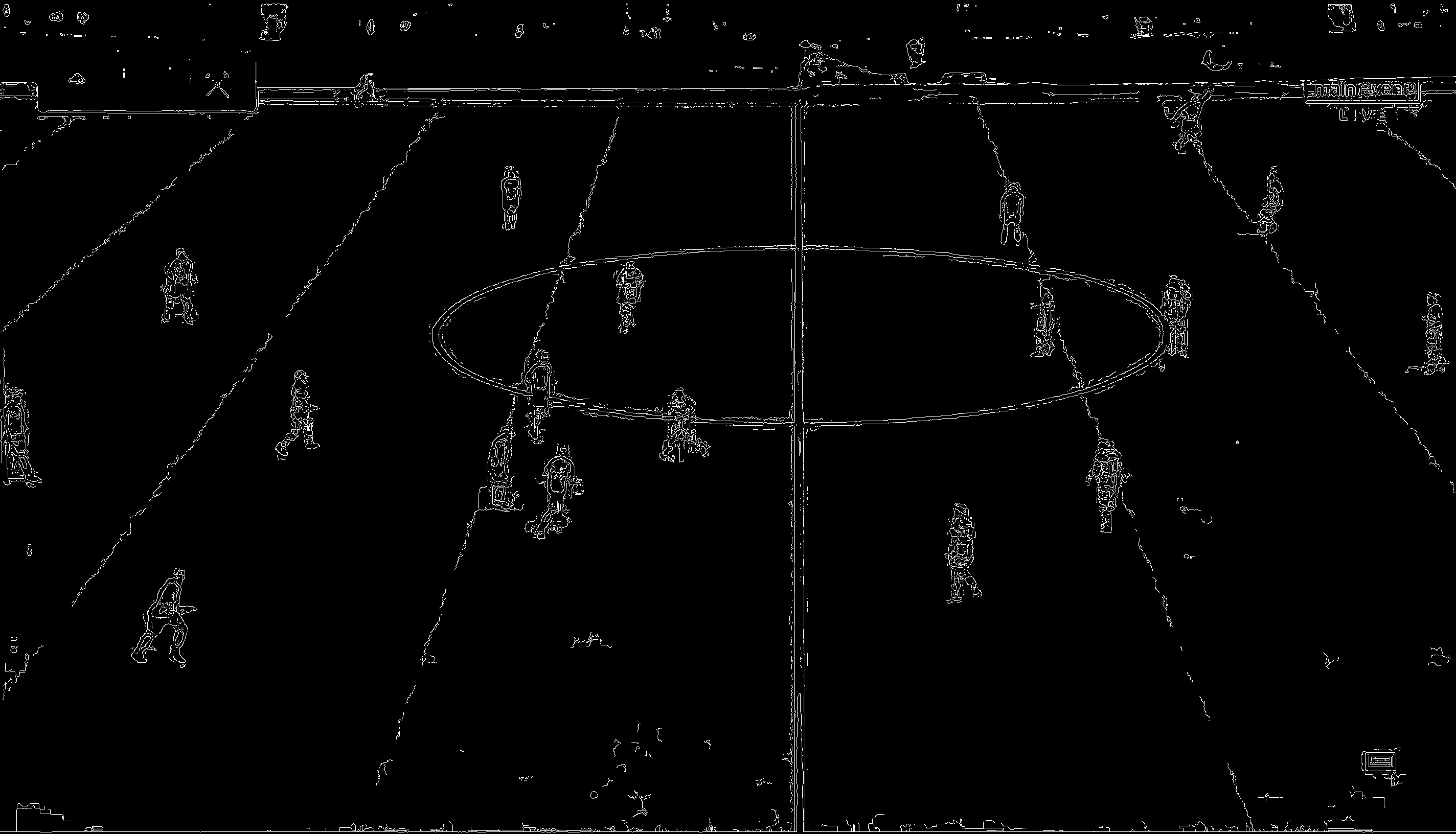}
    \caption{Edge detection after using Canny edge detector}
    \label{fig:gan_before_MC}
\end{figure}

Next part was detecting these lines on this image. The Hough transform, a powerful technique for line detection, effectively converted edge points into lines by transforming them into parameter space. Through parameter tuning including rho, theta, and appropriate thresholds, I achieved accurate line detection on the Soccer field. After obtaining a list of detected lines with their endpoints, we implemented a filtering mechanism based on line angles. I calculated the angles of all the detected lines and only used those which lie between 18 $^{\circ}$ and 89$^{\circ}$. We cannot use 90$^{\circ}$ as the slope calculation blows up. This filtering helped to detect accurate lines on the canny edge detected image. As you can see in the Fig. 6., accurate  lines are detected on the Soccer pitch concluding the Line detection on Soccer pitch.

\begin{figure}[htbp]
    \centering
    \includegraphics[width=0.4\textwidth]{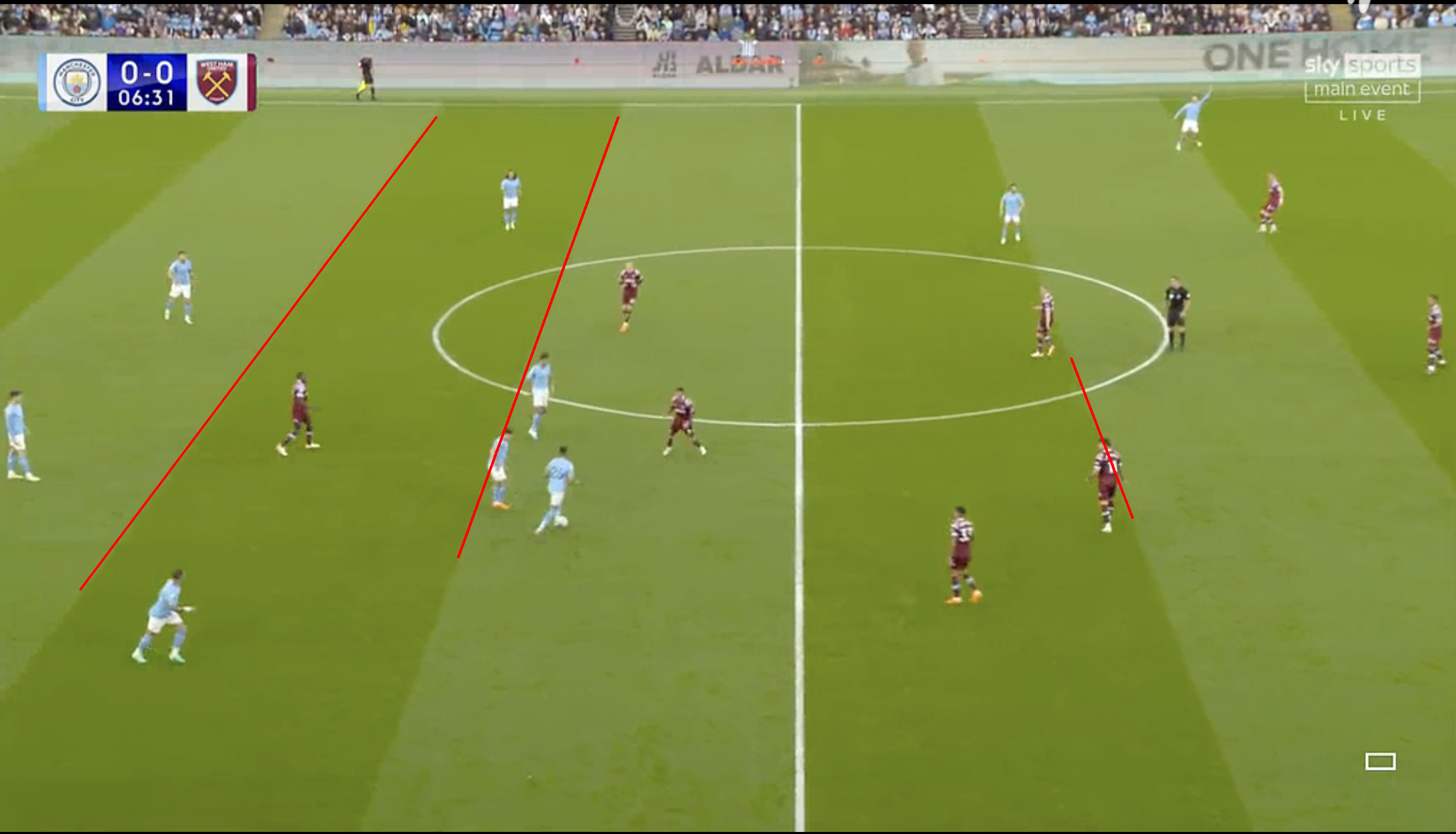}
    \caption{Detected Lines}
    \label{fig:gan_before_MC}
\end{figure}

\subsection{Vanishing point:}
The set of lines we got from Hough detection shows that these lines are not parallel in 2D space. Although, these set of lines are parallel in 3D space. Hence to find the similar parallel lines on the image, calculating Vanishing point becomes necessary. In 2D space, the vanishing point is where parallel lines appear to converge when projected onto a flat surface, creating the illusion of depth. In 3D space, it represents the point at which parallel lines in the three-dimensional world converge when viewed from a specific perspective, aiding in understanding spatial relationships and depth perception in images and scenes.
The algorithm starts by analyzing all lines detected in the image. For each pair of lines (excluding pairs of identical lines), it calculates the intersection point using mathematical principles based on the lines' equations.

Next, the algorithm employs the Random Sample Consensus (RANSAC) technique, a robust estimation method commonly used in computer vision tasks. RANSAC iteratively fits line models through randomly sampled intersection points, aiming to find the best-fitting line model that represents the vanishing point.

During each iteration of RANSAC, two random intersection points are chosen to form a line model. The distances between all intersection points and this line model are computed. Points that lie within a specified threshold distance from the line model are considered inliers.

The algorithm repeats this process for a predetermined number of iterations, adjusting the line model each time to maximize the number of inliers. It keeps track of the line model with the most inliers, indicating a better fit to the actual vanishing point.

The iteration terminates early if a sufficient percentage of points are identified as inliers, providing computational efficiency while ensuring accuracy.

Finally, the vanishing point is determined by averaging the coordinates of the best set of inliers found during the RANSAC process. This average represents the estimated location where parallel lines in the image converge, providing valuable information about the image's perspective and spatial layout. I used a reference code for the RANSAC algorithm from \href{https://medium.com/@KuoyuanLi/detecting-the-vanishing-point-in-one-point-perspective-images-using-computer-vision-algorithms-c4352d4e6c3e}{this blog}.

The output of this is shown in 2 different settings during the game as shown in Fig. 7 and 8.

\begin{figure}[htbp]
    \centering
    \includegraphics[width=0.4\textwidth]{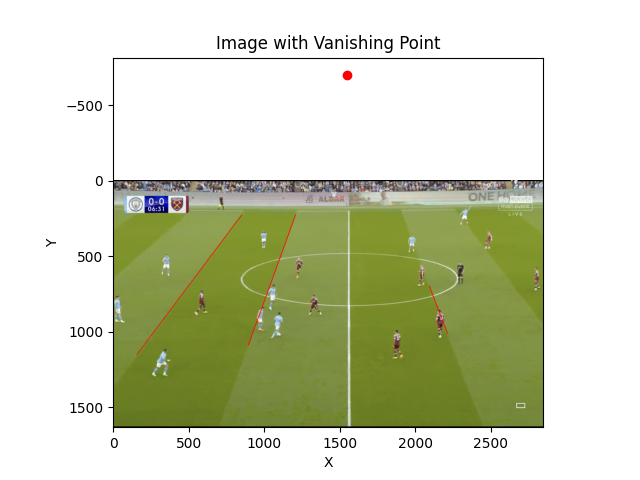}
    \caption{Vanishing point when Camera is looking at the center of the Pitch}
    \label{fig:gan_before_MC}
\end{figure}

\begin{figure}[htbp]
    \centering
    \includegraphics[width=0.4\textwidth]{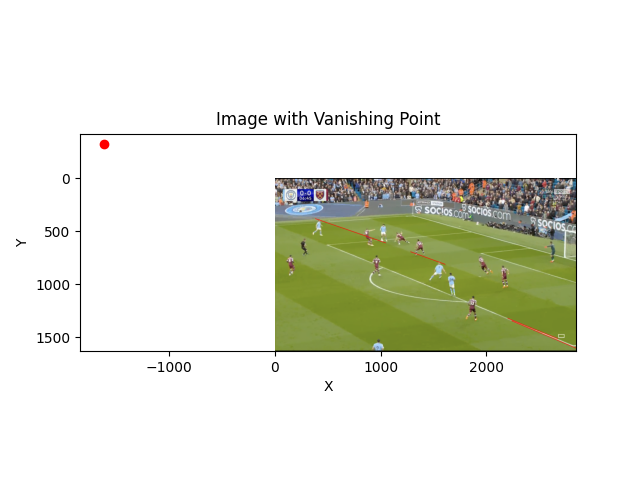}
    \caption{Vanishing point when Camera is looking at one end of the Pitch}
    \label{fig:gan_before_MC}
\end{figure}

This concludes the part of calculating the Vanish point.

\subsection{Keypose Estimation:}
In soccer, the Offside rule, which is crucial for fair play, considers specific body parts like the left and right knees, shoulders, and ankles/toes. Accurate detection of Offside situations relies on keypose estimation, emphasizing the importance of understanding player positions and movements for precise rule enforcement.

For the algorithm of KeyPose detection i defined a PoseEstimator class that uses a pre-trained keypoint detection model to detect and draw keypoints on an input image. The class initializes with parameters for detection quality threshold and keypoint quality threshold. It sets up the model using Torchvision's KeypointRCNN\_ResNet50\_FPN model and necessary transformations. The detect\_and\_draw method takes an image as input, converts it to the appropriate format for the model, performs inference, and draws bounding boxes and keypoints on the image based on the model's predictions. The algorithm to draw keyposes on the input image is given below: \\
1. Convert the input image from BGR to RGB format and normalize the pixel values to the range [0, 1].\\
2. Apply transformations required by the model on the image.\\
3. Perform inference using the pre-trained keypoint detection model.\\
4. Extract scores, keypoints, bounding boxes, and labels from the model's output.\\
5. Iterate through the detected objects and filter based on the detection quality threshold.\\
6. For each detected person (based on the label "person"), draw a bounding box and label on the image.\\
7. Iterate through the keypoints of the detected person, filter based on the keypoint quality threshold, and draw circles at the keypoints of interest (specified by desired\_keypoints).
8. Return the annotated image and a list of collected keypoints.\\

Fig. 9. shows the output of pose estimation and predicting all the necessary keypoints on all the players in the image.

\begin{figure}[htbp]
    \centering
    \includegraphics[width=0.4\textwidth]{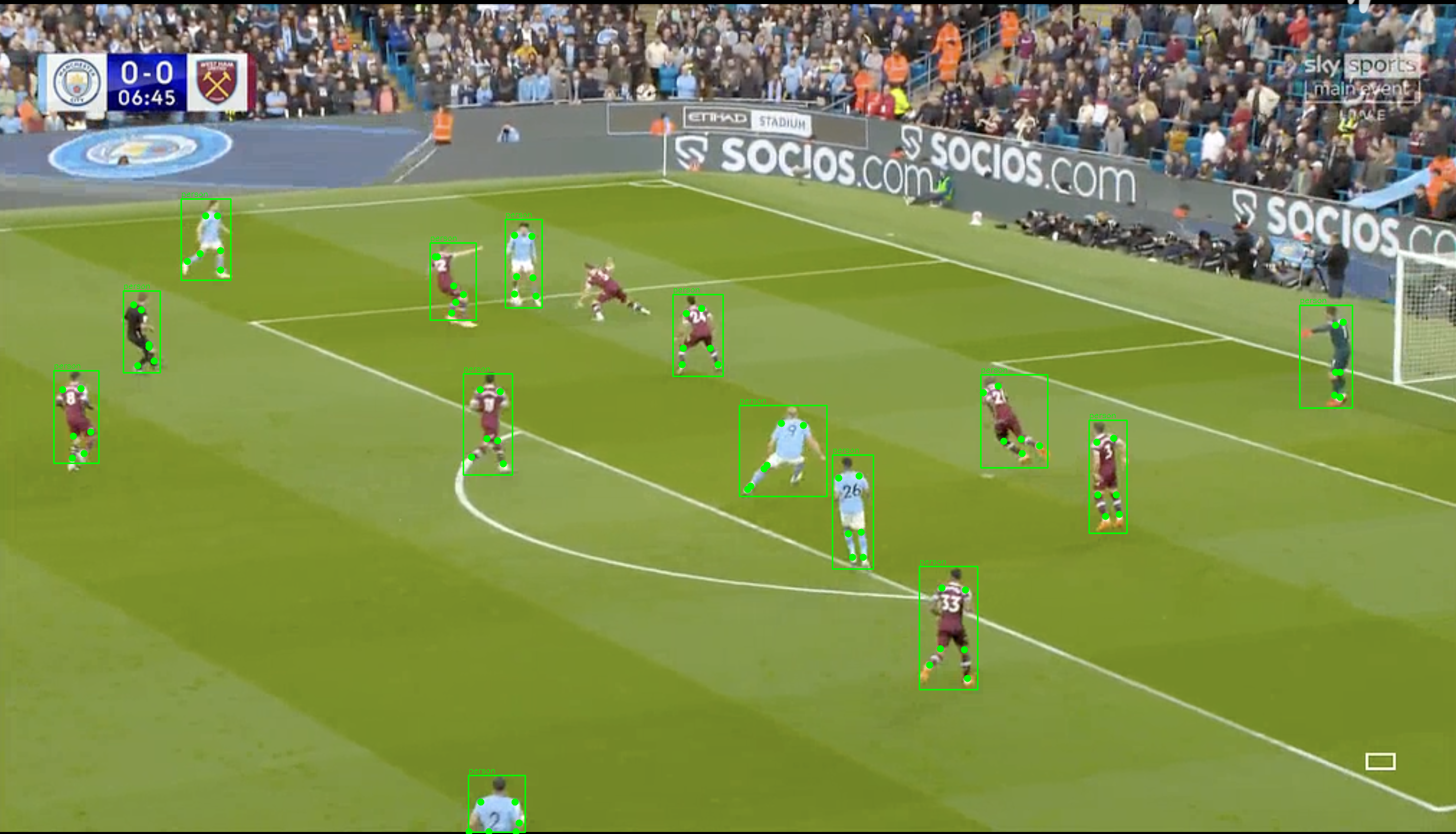}
    \caption{Detected Key points}
    \label{fig:gan_before_MC}
\end{figure}

This concludes the last part of the experiment of predicting key points using pre-trained Keypose estimation model.

\section{Results}

The goal of this project was to draw parallel virtual lines using a single camera view (broadcast view) which pass through the player's key points. These set of lines are potential offside lines and the right most line out of all the virtual lines becomes the actual lines that we need to track during the video for Offside detection. 
As we have a Vanishing point $(x1, y1)$ and a set of key pose points (list of $(x2, y2)$), we draw each and every line passing through all the detected keypoints. Fig. 10. shows the all the lines drawn from Vanishing point $(x1, y1)$.

\begin{figure}[htbp]
    \centering
    \includegraphics[width=0.4\textwidth]{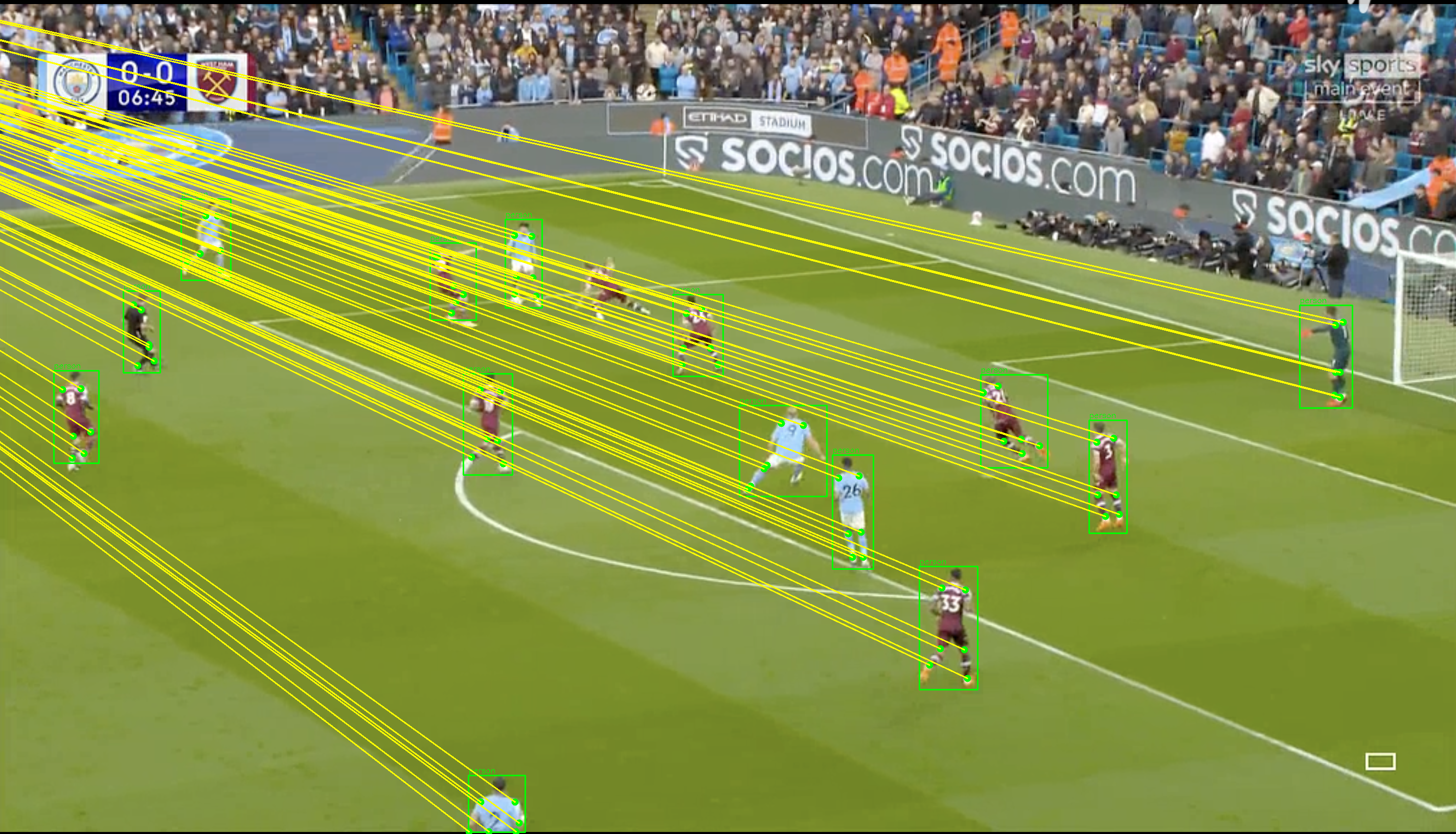}
    \caption{Lines drawn from all the Detected Key points}
    \label{fig:gan_before_MC}
\end{figure}

Although these lines do not look like they are parallel in 2D space, these lines are parallel in 3D space. We now need the lines going through the last defender of the defending team which are the right most set of lines. Note that in the given Fig. 10. the right most lines pass through the Goalkeeper of the defending team, but Goalkeeper is not involved in the offside rule (barring a rare edge case which is out of scope of this project). Fig. 11. shows the final goal of this project achieved by getting the set of lines of the last defender of the defending team.

\begin{figure}[htbp]
    \centering
    \includegraphics[width=0.4\textwidth]{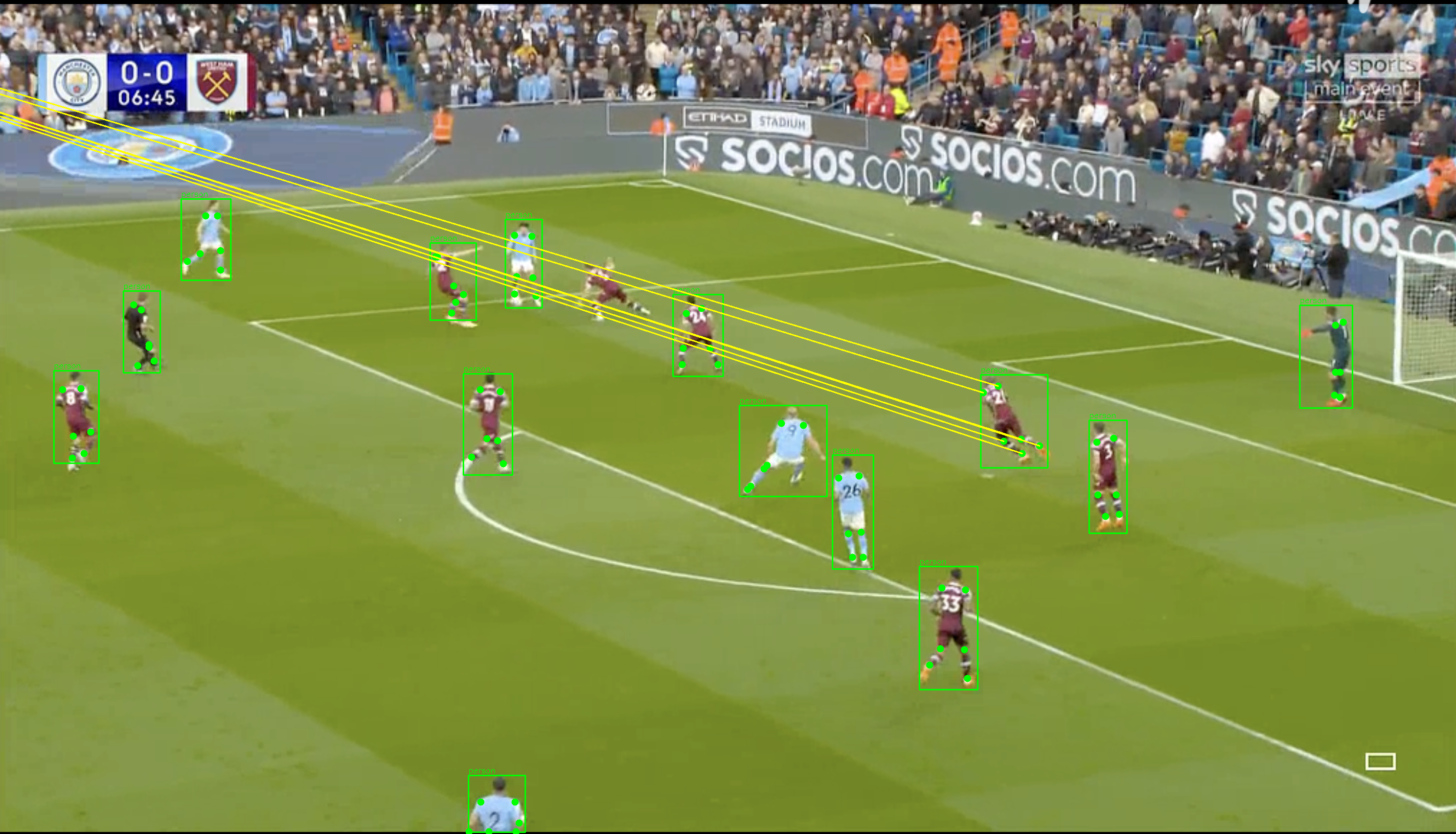}
    \caption{Lines drawn from all the Detected Key points of the Last defender (right most defender)}
    \label{fig:gan_before_MC}
\end{figure}

The right most line from Fig. 11. is the line to check offside in real time video which is the final objective of this project. I do this by calculating the slopes of all the required lines an using the minimum absolute value of the slope always giving the final line for offside.

\section{Discussion}
Leveraging basic image processing techniques involves utilizing fundamental methods to analyze and manipulate images effectively. These techniques are essential for various tasks in computer vision and image analysis. Here's an elaboration on the mentioned techniques:
\subsubsection{Different Image Spaces}Images can be represented in different color spaces like RGB, HSV, LAB, etc. Each color space has its advantages for specific tasks. For instance, HSV is useful for segmenting objects based on color, while LAB is beneficial for color correction and analyzing image gradients.
\subsubsection{Canny Edge Detector}The Canny edge detector is a popular technique for detecting edges in images. It works by identifying areas of rapid intensity change, which often correspond to object boundaries or significant features in the image.
\subsubsection{Hough Lines Transform}The Hough transform is used to detect lines in an image, which is valuable for tasks like line detection, lane detection in autonomous vehicles, and shape recognition. The Hough Lines Transform, a variant of the Hough transform, detects lines using polar coordinates and is robust to noise.
\subsubsection{Keypose Estimation using PyTorch}Keypose estimation involves identifying key points or landmarks in an image, such as joints in a human body. PyTorch, a popular deep learning framework, provides pre-trained models and tools for keypoint detection, making it easier to implement complex computer vision tasks like pose estimation.
\subsubsection{Calculating Vanishing Points}Vanishing points are critical in computer vision for understanding the 3D structure of a scene from 2D images. By analyzing the convergence of parallel lines in an image, vanishing points can be calculated, aiding in tasks like camera calibration, depth estimation, and scene understanding.

\section{Future Work}
Tracking players based on their jersey/kit colors and determining the attacking player and defending player using KLT (Kanade-Lucas-Tomasi) tracker. \cite{muthuraman2018vision} used a KLT tracker to track the players based on the jersey colors using a UI interface which can be bypassed as well. Detecting the referees and goalkeeper in the game in every frame, as they are not a part of the Offside (barring the rare edge cases such as goalkeeper is ahead of the last defender of the team).
Detecting the “last” line of the defender of a team and attacker of the other team and then labelling if the player of attacking team is offside or not. Along with this, the pretrained model KeypointRCNN\_ResNet50 is accurate but slow. Slower models would make the video processing slow. Hence, finding an accurate and faster key pose estimator can work well on a real-time video.

\section{Conclusion}
I came up with a novel idea of Offside detection to achieve highest accuracy in finding the offside line. But the question arises why Single Camera should be used ? The current SOTA (state of the art) for offside detection instilled as Video Assistant Referee (VAR), uses 30+ cameras around the stadium which makes it a huge investment for all the teams/leagues. Since this is a huge investment, only the rich clubs/teams/stadiums/leagues can afford this technology. The low budget teams cannot afford such tech as they hardly use more than 2 cameras making the sport unfair for the low-budget teams.Bringing in this technology makes the game fair for all sort of financially backed clubs/teams/stadiums/leagues. Although, SOTA tech is highly accurate, there is a debate among the people regarding its use making the sport monotonous and taking out the “uncertainty factor” which many people like about any sport. A minor error or benefit of doubt makes the game more lively. Hence, while the research area in academia leverages expensive technology to get the highest accuracy possible, cutting down on cost significantly at highest accuracy possible using the principles of Augmented Reality by making the Sport of Soccer fair is an inspiration to implement such a project and take this research forward.

\section*{Acknowledgment}

I wish to really thank my advisor Dr. Kinsman who guided me through all the problems I faced during the course of this research project. The idea of using the light green and dark green stripes on the Soccer pitch was inspired by Dr. Kinsman's quote, "As an engineer, we can use all sorts of unfair advantages we have".



\bibliographystyle{IEEEtran}
\bibliography{sample}
%



\end{document}